\documentclass[11pt,a4paper]{article}
\usepackage[hyperref]{acl2019}
\usepackage{times}
\usepackage{latexsym}

\usepackage{url}
\usepackage{amsmath}
\usepackage{graphicx}
\usepackage[normalem]{ulem}
\usepackage{subcaption}
\usepackage{booktabs}

\usepackage{outlines}  %

\usepackage{todonotes}

\aclfinalcopy %

\title{BERT Rediscovers the Classical NLP Pipeline}

\author{
Ian Tenney$^1$ \quad Dipanjan Das$^1$ \quad Ellie Pavlick$^{1,2}$ \\
$^1$Google Research \quad $^2$Brown University \\
\texttt{\{iftenney,dipanjand,epavlick\}@google.com} \\
}

\date{}

\begin{document}

\maketitle
\begin{abstract}
Pre-trained text encoders have rapidly advanced the state of the art on many NLP tasks. We focus on one such model, BERT, and aim to quantify where linguistic information is captured within the network. We find that the model represents the steps of the traditional NLP pipeline in an interpretable and localizable way, and that the regions responsible for each step appear in the expected sequence: POS tagging, parsing, NER, semantic roles, then coreference. Qualitative analysis reveals that the model can and often does adjust this pipeline dynamically, revising lower-level decisions on the basis of disambiguating information from higher-level representations.
\end{abstract}

\section{Introduction}
\label{sec:intro}

Pre-trained sentence encoders such as ELMo \citep{peters2018deep} and BERT \citep{devlin2018bert} have rapidly improved the state of the art on many NLP tasks, and seem poised to displace both static word embeddings \citep{mikolov2013distributed} and discrete pipelines \citep{manning2014stanford} as the basis for natural language processing systems. While this has been a boon for performance, it has come at the cost of interpretability, and it remains unclear whether such models are in fact learning the kind of abstractions that we intuitively believe are important for representing natural language, or are simply modeling complex co-occurrence statistics.

A wave of recent work has begun to ``probe'' state-of-the-art models to understand whether they are representing language in a satisfying way. Much of this work is behavior-based, designing controlled test sets and analyzing errors in order to reverse-engineer the types of abstractions the model may or may not be representing \citep[e.g.][]{conneau2018cram,marvin2018targeted,poliak2018collecting}. Parallel efforts inspect the structure of the network directly, to assess whether there exist localizable regions associated with distinct types of linguistic decisions.
Such work has produced evidence that deep language models can encode a range of syntactic and semantic information \citep[e.g.][]{shi2016does,belinkov2018thesis,tenney2018what}, and that more complex structures are represented hierarchically in the higher layers of the model \citep{peters2018dissecting,blevins2018hierarchical}.

We build on this latter line of work, focusing on the BERT model \citep{devlin2018bert}, and use a suite of probing tasks \citep{tenney2018what} derived from the traditional NLP pipeline to quantify where specific types of linguistic information are encoded. 
Building on observations \citep{peters2018dissecting} that lower layers of a language model encode more local syntax while higher layers capture more complex semantics, we present two novel contributions.
First, we present an analysis that spans the common components of a traditional NLP pipeline. We show that the order in which specific abstractions are encoded reflects the traditional hierarchy of these tasks.
Second, we qualitatively analyze how individual sentences are processed by the BERT network, layer-by-layer. We show that while the pipeline order holds in aggregate, the model can allow individual decisions to depend on each other in arbitrary ways, deferring ambiguous decisions or revising incorrect ones based on higher-level information.

\section{Model}
\label{sec:experimental-model}

\paragraph{Edge Probing.}
Our experiments are based on the ``edge probing'' approach of \citet{tenney2018what}, which aims to measure how well information about linguistic structure can be extracted from a pre-trained encoder. Edge probing decomposes structured-prediction tasks into a common format, where a probing classifier receives spans $\mathtt{s}_1 = [i_1, j_1)$ and (optionally) $\mathtt{s}_2 = [i_2, j_2)$ and must predict a label such as a constituent or relation type.\footnote{For single-span tasks (POS, entities, and constituents), $\mathtt{s}_2$ is not used. For POS, $\mathtt{s}_1 = [i,i+1)$ is a single token.}
The probing classifier has access only to the per-token contextual vectors \textit{within} the target spans, and so must rely on the encoder to provide information about the relation between these spans and their role in the sentence.

We use eight
labeling tasks from the edge probing suite: part-of-speech (POS), constituents (Consts.), dependencies (Deps.), entities, semantic role labeling (SRL), coreference (Coref.), semantic proto-roles \citep[SPR;][]{reisinger2015semantic}, and relation classification (SemEval). These tasks are derived from standard benchmark datasets, and evaluated with a common metric--micro-averaged F1--to facilitate comparison across tasks.
\footnote{We use the code from \url{https://github.com/jsalt18-sentence-repl/jiant}. Dependencies is the English Web Treebank \citep{silveira14gold}, SPR is the SPR1 dataset of \citep{teichert2017semantic}, and relations is SemEval 2010 Task 8 \citep{hendrickx2009semeval}. All other tasks are from OntoNotes 5.0 \citep{weischedel2013ontonotes}.}

\paragraph{BERT.}
The BERT model \citep{devlin2018bert} has shown state-of-the-art performance on many tasks, and its deep Transformer architecture \citep{vaswani2017attention} is typical of many recent models \citep[e.g.][]{radford2018improving,radford2019language,liu2019multi}.
We focus on the stock BERT models (base and large, uncased), which are trained with a multi-task objective (masked language modeling and next-sentence prediction) over a 3.3B word English corpus.
Since we want to understand how the network represents language as a result of pretraining, we follow \citet{tenney2018what} (departing from standard BERT usage) and freeze the encoder weights.
This prevents the encoder from re-arranging its internal representations to better suit the probing task.

Given input tokens
$T = [t_0, t_1, \ldots, t_n]$, a deep encoder produces a set of layer activations $H^{(0)}, H^{(1)}, \ldots, H^{(L)}$, where $H^{(\ell)} = [\mathbf{h}_0^{(\ell)}, \mathbf{h}_1^{(\ell)}, \ldots, \mathbf{h}_n^{(\ell)}]$ are the activation vectors of the $\ell^{th}$ encoder layer and $H^{(0)}$ corresponds to the non-contextual word(piece) embeddings. We use a weighted sum across layers (\S\ref{sec:scalar-mixing}) to pool these into a single set of per-token representation vectors $H = [\mathbf{h}_0, \mathbf{h}_1, \ldots, \mathbf{h}_n]$, and train a probing classifier $P_{\tau}$ for each task using the architecture and procedure of \citet{tenney2018what}.

\paragraph{Limitations}
This work is intended to be exploratory. We focus on one particular encoder--BERT--to explore how information can be organized in a deep language model, and further work is required to determine to what extent the trends hold in general.
Furthermore, our work carries the limitations of all inspection-based probing: the fact that a linguistic pattern is not observed by our probing classifier does not guarantee that it is not there, and the observation of a pattern does not tell us how it is used. For this reason, we emphasize the importance of combining structural analysis with behavioral studies (as discussed in \S~\ref{sec:intro}) to provide a more complete picture of what information these models encode and how that information affects performance on downstream tasks.

\section{Metrics}
\label{sec:metrics}

We define two complementary metrics.
The first, scalar mixing weights (\S\ref{sec:scalar-mixing}) tell us which layers, in combination, are most relevant when a probing classifier has access to the whole BERT model. The second, cumulative scoring (\S\ref{sec:cumulative-scoring}) tells us how much higher we can score on a probing task with the introduction of each layer. These metrics provide complementary views on what is happening inside the model. Mixing weights are learned solely from the training data--they tell us which layers the probing model finds most useful. In contrast, cumulative scoring is derived entirely from an evaluation set, and tell us how many layers are needed for a correct prediction.

\subsection{Scalar Mixing Weights}
\label{sec:scalar-mixing}
To pool across layers, we use the scalar mixing technique introduced by the ELMo model. Following Equation~(1) of \citet{peters2018deep}, for each task we introduce scalar parameters $\gamma_{\tau}$ and $a^{(0)}_{\tau}, a^{(1)}_{\tau}, \ldots, a^{(L)}_{\tau}$, and let:
\begin{equation}\label{eq:mixing-weights}
    \mathbf{h}_{i,\tau} = \gamma_{\tau} \sum_{\ell = 0}^L s^{(\ell)}_{\tau} \mathbf{h}_i^{(\ell)}
\end{equation}
where $\mathbf{s}_{\tau} = \mathrm{softmax}(\mathbf{a}_{\tau})$. We learn these weights jointly with the probing classifier ${P}_{\tau}$, in order to allow it to extract information from the many layers of an encoder without adding a large number of parameters. After the probing model is trained, we extract the learned coefficients in order to estimate the contribution of different layers to that particular task. We interpret higher weights as evidence that the corresponding layer contains more information related to that particular task.

\paragraph{Center-of-Gravity.}
As a summary statistic, we define the mixing weight center of gravity as:
\begin{equation}\label{eq:mixing-cog}
    \bar{E}_{s}[\ell] = \sum_{\ell=0}^L \ell \cdot s^{(\ell)}_{\tau}
\end{equation}
\noindent This reflects the average layer attended to for each task; intuitively, we can interpret a higher value to mean that the information needed for that task is captured by higher layers.

\begin{figure}[t]
    \centering
    \includegraphics[width=\linewidth]{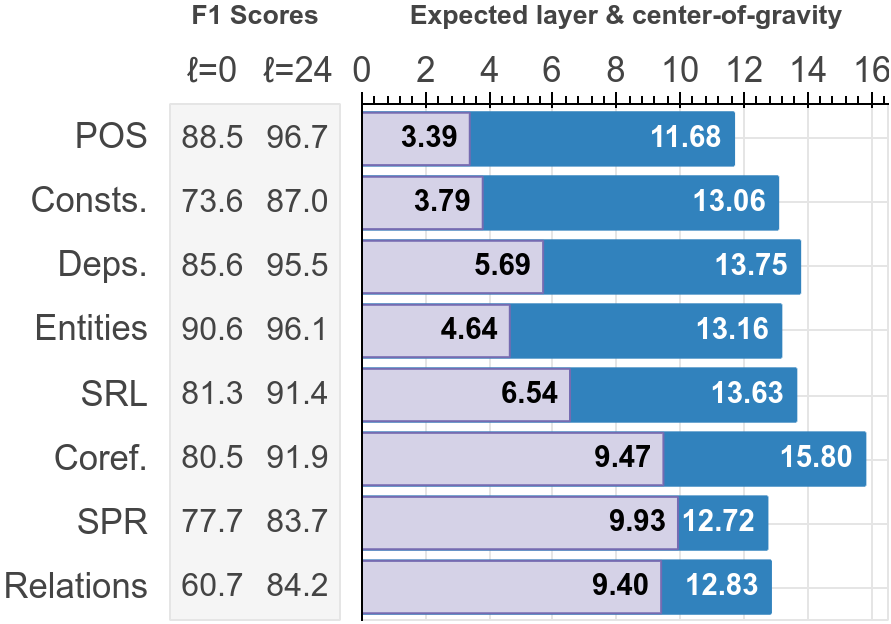}
    \caption{Summary statistics on BERT-large. Columns on left show F1 dev-set scores for the baseline (${P}^{(0)}_{\tau}$) and full-model (${P}^{(L)}_{\tau}$) probes. Dark (blue) are the mixing weight center of gravity (Eq.~\ref{eq:mixing-cog}); light (purple) are the expected layer from the cumulative scores (Eq.~\ref{eq:exp-layer}).}
    \label{fig:summary-stats}
\end{figure}

\begin{figure}[t!]
  \centering
  \includegraphics[width=\linewidth]{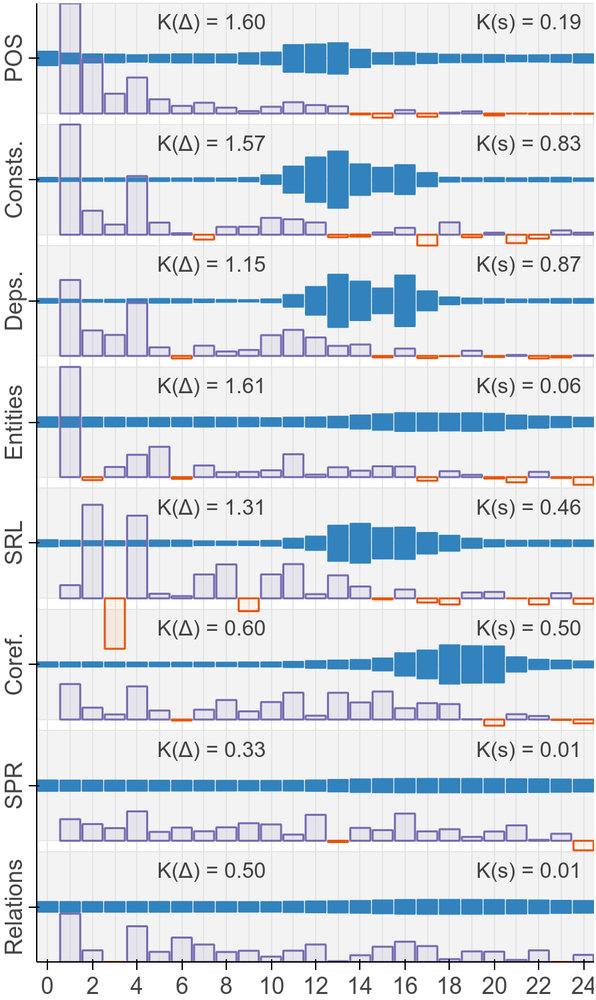}
  \caption{Layer-wise metrics on BERT-large. Solid (blue) are mixing weights $s^{(\ell)}_{\tau}$ (\S\ref{sec:scalar-mixing}); outlined (purple) are differential scores $\Delta^{(\ell)}_{\tau}$ (\S\ref{sec:cumulative-scoring}), normalized for each task. Horizontal axis is encoder layer.}
 \label{fig:score-distro}
\end{figure}

\subsection{Cumulative Scoring}
\label{sec:cumulative-scoring}

We would like to estimate at which layer in the encoder a target $(\mathtt{s}_1, \mathtt{s}_2, \mathtt{label})$ can be correctly predicted.
Mixing weights cannot tell us this directly, because they are learned as parameters and do not correspond to a distribution over data. 
A naive classifier at a single layer cannot either, because information about a particular span may be spread out across several layers, and as observed in \citet{peters2018dissecting} the encoder may choose to discard information at higher layers.

To address this, we train a series of classifiers $\{{P}^{(\ell)}_{\tau}\}_\ell$ which use scalar mixing (Eq.~\ref{eq:mixing-weights}) to attend to layer $\ell$ as well as \textit{all previous} layers. ${P}^{(0)}_{\tau}$ corresponds to a non-contextual baseline that uses only a bag of word(piece) embeddings, while ${P}^{(L)}_{\tau} = {P}_{\tau}$ corresponds to probing all layers of the BERT model.

These classifiers are cumulative, in the sense that $P^{(\ell+1)}_{\tau}$ has a similar number of parameters but with access to strictly more information than $P^{(\ell)}_{\tau}$, and we see intuitively that performance (F1 score) generally increases as more layers are added.\footnote{Note that if a new layer provides distracting features, the probing model can overfit and performance can drop. We see this in particular in the last 1-2 layers (Figure~\ref{fig:score-distro}).}
We can then compute a differential score $\Delta^{(\ell)}_{\tau}$, which measures how much better we do on the probing task if we observe one additional encoder layer $\ell$:
\begin{equation}\label{eq:differential-score}
    \Delta^{(\ell)}_{\tau} = \mathrm{Score}({P}^{(\ell)}_{\tau}) - \mathrm{Score}({P}^{(\ell-1)}_{\tau})
\end{equation}

\paragraph{Expected Layer.}
Again, we compute a (pseudo)\footnote{This is not a true expectation because the F1 score is not an expectation over examples.} expectation over the differential scores as a summary statistic. To focus on the behavior of the contextual encoder layers, we omit the contribution of both the ``trivial'' examples resolved at layer $0$, as well as the remaining headroom from the full model. Let:
\begin{equation}\label{eq:exp-layer}
    \bar{E}_{\Delta}[\ell] = \frac{\sum_{\ell = 1}^L \ell \cdot \Delta^{(\ell)}_{\tau}}{\sum_{\ell = 1}^L \Delta^{(\ell)}_{\tau}} 
\end{equation}
This can be thought of as, approximately, the expected layer at which the probing model correctly labels an example, assuming that example is resolved at \textit{some} layer $\ell \ge 1$ of the encoder.

\section{Results}
\label{sec:results}

Figure~\ref{fig:summary-stats} reports summary statistics and absolute F1 scores, and Figure~\ref{fig:score-distro} reports per-layer metrics. Both show results on the 24-layer BERT-large model. We also report $K(\star) = \mathrm{KL}(\star || \mathrm{Uniform})$ to estimate how non-uniform\footnote{$\mathrm{KL}(\star || \mathrm{Uniform}) = -H(\star) + \mathrm{Constant}$, so higher values correspond to lower entropy.} each statistic ($\star = s_{\tau}, \Delta_{\tau}$) is for each task.

\paragraph{Linguistic Patterns.}
We observe a consistent trend across both of our metrics, with the tasks encoded in a natural progression: POS tags processed earliest, followed by constituents, dependencies, semantic roles, and coreference. That is, it appears that basic syntactic information appears earlier in the network, while high-level semantic information appears at higher layers. We note that this finding is consistent with initial observations by \citet{peters2018dissecting}, which found that constituents are represented earlier than coreference.

In addition, we observe that in general, syntactic information is more localizable, with weights related to syntactic tasks tending to be concentrated on a few layers (high $K(s)$ and $K(\Delta)$), while information related to semantic tasks is generally spread across the entire network. 
For example, we find that for semantic relations and proto-roles (SPR), the mixing weights are close to uniform, and that nontrivial examples for these tasks are resolved gradually across nearly all layers. For entity labeling many examples are resolved in layer 1, but with a long tail thereafter, and only a weak concentration of mixing weights in high layers. Further study is needed to determine whether this is because BERT has difficulty representing the correct abstraction for these tasks, or because semantic information is inherently harder to localize.

\paragraph{Comparison of Metrics.}
For many tasks, we find that the differential scores are highest in the first few layers of the model (layers 1-7 for BERT-large), i.e. most examples can be correctly classified very early on. We attribute this to the availability of heuristic shortcuts: while challenging examples may not be resolved until much later, many cases can be guessed from shallow statistics.
Conversely, we observe that the learned mixing weights are concentrated much later, layers 9-20 for BERT-large.
We observe--particularly when weights are highly concentrated--that the highest weights are found on or just after the \textit{highest} layers which give an improvement $\Delta^{(\ell)}_{\tau}$ in F1 score for that task.

This helps explain the observations on the semantic relations and SPR tasks: cumulative scoring shows continued improvement up to the highest layers of the model, while the lack of concentration in the mixing weights suggest that the BERT encoder does not expose a localized set of features that encode these more semantic phenomena. Similarly for entity types, we see continued improvements in the higher layers -- perhaps related to fine-grained semantic distinctions like "Organization" (\texttt{ORG}) vs. "Geopolitical Entity" (\texttt{GPE}) -- while the low value for the \textit{expected} layer reflects that many examples require only limited context to resolve.

\paragraph{Comparison of Encoders.}
We observe the same general ordering on the 12-layer BERT-base model (Figure~\ref{fig:supp-summary-stats}). In particular, there appears to be a ``stretching effect'', where the representations for a given task tend to concentrate at the same layers \textit{relative to the top of the model}; this is illustrated side-by-side in Figure~\ref{fig:supp-score-distro}.

\subsection{Per-Example Analysis}
\label{sec:per-example}

\begin{figure}[t!]
  \centering
  \begin{subfigure}[b]{\linewidth}
    \centering
    \caption{he smoked \textbf{toronto} in the playoffs with six hits, seven walks and eight stolen bases ...}
    \includegraphics[width=\linewidth]{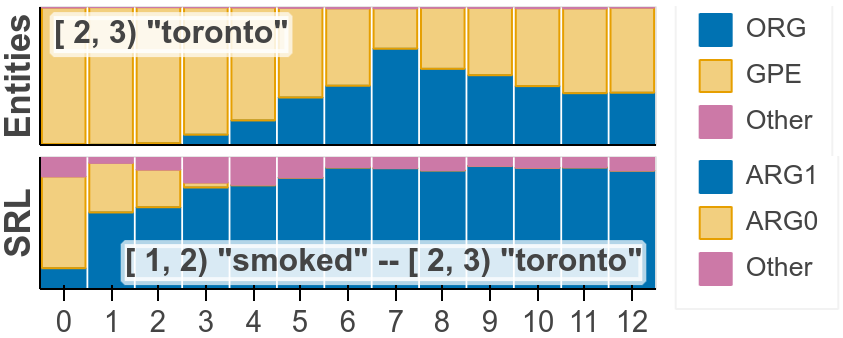}
  \end{subfigure}
  ~\\
  \begin{subfigure}[b]{\linewidth}
    \centering
    \caption{china \textbf{today} blacked out a cnn interview that was ...}
    \includegraphics[width=\linewidth]{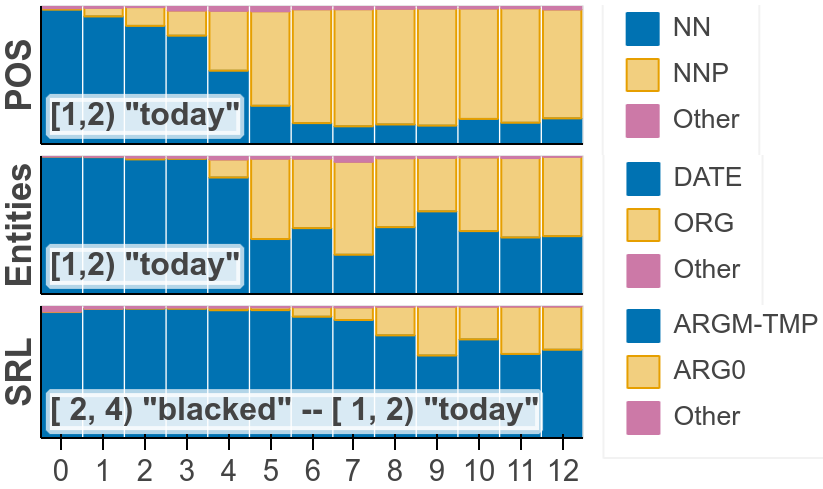}
  \end{subfigure}
  \caption{Probing classifier predictions across layers of BERT-base. Blue is the correct label; orange is the incorrect label with highest average score over layers. Bar heights are (normalized) probabilities ${P}^{(\ell)}_{\tau}(\mathtt{label}|\mathtt{s}_1, s_2)$. In the interest of space, only selected annotations are shown.}
  \label{fig:example-traces}
\end{figure}

We explore, qualitatively, how beliefs about the structure of individual sentences develop over the layers of the BERT network. The OntoNotes development set contains annotations for five of our probing tasks: POS, constituents, entities, SRL, and coreference. We compile the predictions of the per-layer classifiers ${P}^{(\ell)}_{\tau}$ for each task. Because many annotations are uninteresting -- for example, 89\% of part-of-speech tags are correct at layer 0 -- we use a heuristic to identify ambiguous sentences to visualize.\footnote{Specifically, we look for target edges ($\mathtt{s}_1$, $\mathtt{s}_2$, $\mathtt{label}$) where the highest scoring label has an average score $\frac{1}{L+1} \sum_{\ell = 0}^{L} P_{\tau}^{(\ell)}(\mathtt{label}|\mathtt{s}_1,\mathtt{s}_2) \le 0.7$, and look at sentences with more than one such edge.}
Figure~\ref{fig:example-traces} shows two selected examples, and more are presented in Appendix~\ref{sec:appendix-examples}.

We find that while the pipeline order holds on average (Figure~\ref{fig:score-distro}), for individual examples the model is free to and often does choose a different order. In the first example, the model originally (incorrectly) assumes that \textit{``Toronto''} refers to the city, tagging it as a \texttt{GPE}. However, after resolving the semantic role -- determining that \textit{``Toronto''} is the thing getting \textit{``smoked''} (\texttt{ARG1}) -- the entity-typing decision is revised in favor of \texttt{ORG} (i.e. the sports team). In the second example, the model initially tags \textit{``today''} as a common noun, date, and temporal modifier (\texttt{ARGM-TMP}). However, this phrase is ambiguous, and it later reinterprets \textit{``china today''} as a proper noun (i.e. the TV network) and updates its beliefs about the entity type (to \texttt{ORG}), followed by the semantic role (reinterpreting it as the agent \texttt{ARG0}).

\section{Conclusion}
\label{sec:conclusion}

We employ the edge probing task suite to explore how the different layers of the BERT network can resolve syntactic and semantic structure within a sentence. We present two complementary measurements: scalar mixing weights, learned from a training corpus, and cumulative scoring, measured on an evaluation set, and show that a consistent ordering emerges. 
We find that while this traditional pipeline order holds in the aggregate, on individual examples the network can resolve out-of-order, using high-level information like predicate-argument relations to help disambiguate low-level decisions like part-of-speech. 
This provides new evidence corroborating that deep language models can represent the types of syntactic and semantic abstractions traditionally believed necessary for language processing, and moreover that they can model complex interactions between different levels of hierarchical information.

\section*{Acknowledgments}
Thanks to Kenton Lee, Emily Pitler, and Jon Clark for helpful comments and feedback, and to the members of the Google AI Language team for many productive discussions.

\bibliography{main}
\bibliographystyle{acl_natbib}

\clearpage

\appendix
\section{Appendix}
\label{sec:appendix}

\renewcommand{\thefigure}{A.\arabic{figure}}
\setcounter{figure}{0}

\subsection{Comparison of Encoders}
\label{sec:appendix-comparison}
We reproduce Figure~1 and Figure~2 (which depict metrics on BERT-large) from the main paper below, and show analogous plots for the BERT-base models. We observe that the most important layers for a given task appear in roughly the same \textit{relative} position on both the 24-layer BERT-large and 12-layer BERT-base models, and that tasks generally appear in the same order.

\begin{figure}[ht!]
  \centering
  \includegraphics[width=0.50\linewidth]{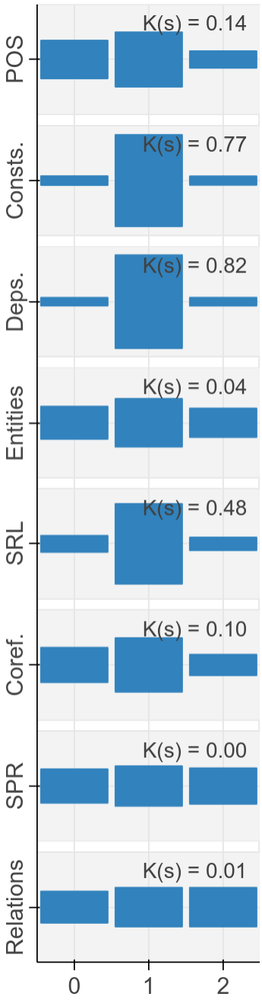}
  \caption{Scalar mixing weights for the ELMo encoder. Layer 0 is the character CNN that produces per-word representations, and layers 1 and 2 are the LSTM layers.}
  \label{fig:supp-elmo-weights}
\end{figure}

Additionally, in Figure~\ref{fig:supp-elmo-weights} we show scalar mixing weights for the ELMo encoder \citep{peters2018deep}, which consists of two LSTM layers over a per-word character CNN. We observe that the first LSTM layer (layer 1) is most informative for all tasks, which corroborates the observations of Figure~2 of \citet{peters2018deep}. As with BERT, we observe that the weights are only weakly concentrated for the relations and SPR tasks. However, unlike BERT, we see only a weak concentration in the weights on the coreference task, which agrees with the finding of \citet{tenney2018what} that ELMo presents only weak features for coreference.

\begin{figure*}[t]
    \centering
  \begin{subfigure}[b]{0.49\linewidth}
    \centering
    \includegraphics[width=\linewidth]{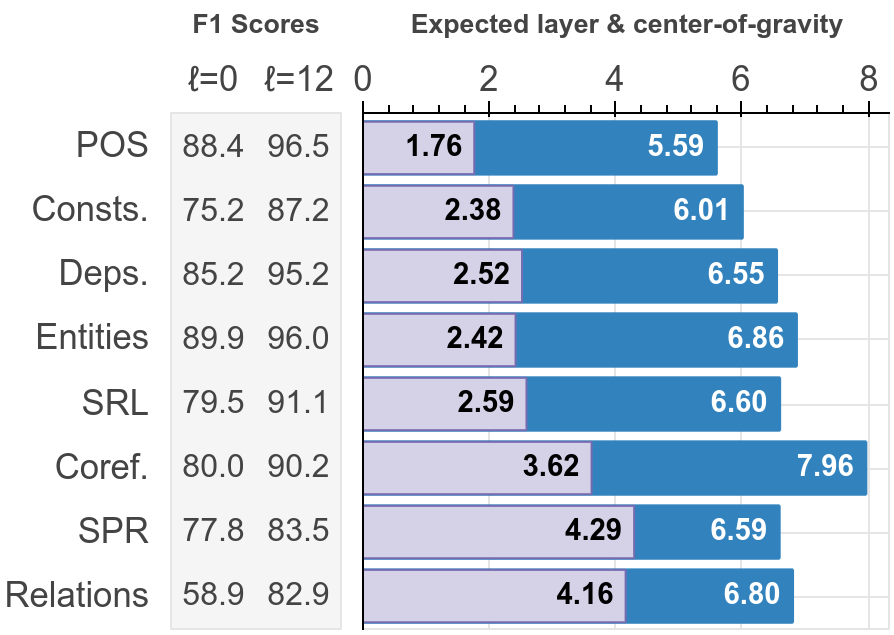}
    \caption{BERT-base}
  \end{subfigure}
  ~
  \begin{subfigure}[b]{0.49\linewidth}
    \centering
    \includegraphics[width=\linewidth]{figures/summary_stats_stacked}
    \caption{BERT-large}
  \end{subfigure}
  \caption{Summary statistics on BERT-base (left) and BERT-large (right). Columns on left show F1 dev-set scores for the baseline (${P}^{(0)}_{\tau}$) and full-model (${P}^{(L)}_{\tau}$) probes. Dark (blue) are the mixing weight center of gravity; light (purple) are the expected layer from the cumulative scores.}
  \label{fig:supp-summary-stats}
\end{figure*}

\begin{figure*}[t]
  \centering
  \begin{subfigure}[b]{0.49\linewidth}
    \centering
    \includegraphics[width=\linewidth]{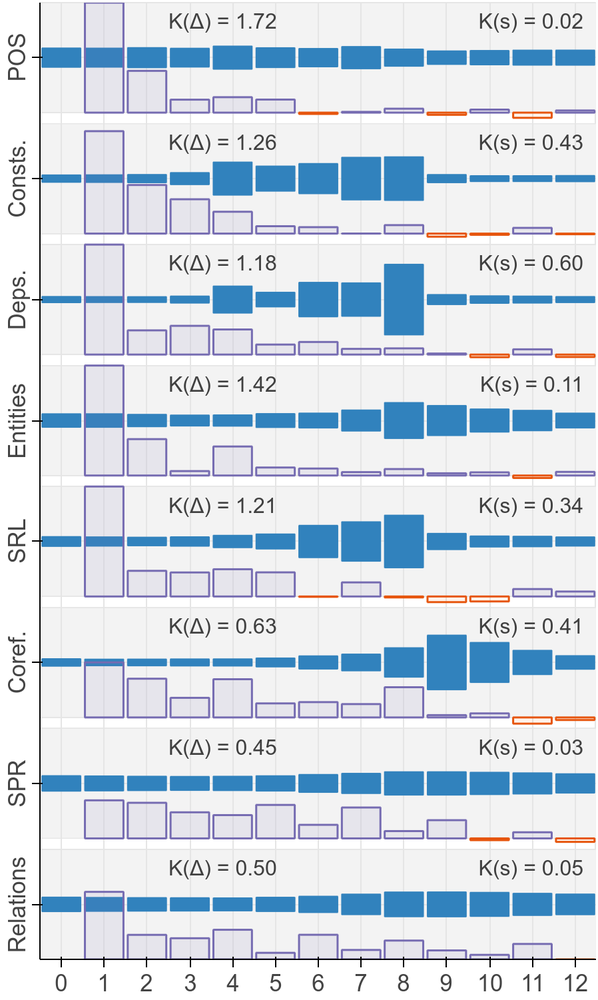}
    \caption{BERT-base}
  \end{subfigure}
  ~
  \begin{subfigure}[b]{0.49\linewidth}
    \centering
    \includegraphics[width=\linewidth]{figures/bylayer_large_compressed}
    \caption{BERT-large}
  \end{subfigure}
  \caption{Layer-wise metrics on BERT-base (left) and BERT-large (right). Solid (blue) are mixing weights $s^{(\ell)}_{\tau}$; outlined (purple) are differential scores $\Delta^{(\ell)}_{\tau}$, normalized for each task. Horizontal axis is encoder layer.}
 \label{fig:supp-score-distro}
\end{figure*}

\subsection{Additional Examples}
\label{sec:appendix-examples}
We provide additional examples in the style of Figure~\ref{fig:example-traces}, which illustrate sequential decisions in the layers of the BERT-base model.

\begin{figure}[ht!]
    \centering
    \includegraphics[width=\linewidth]{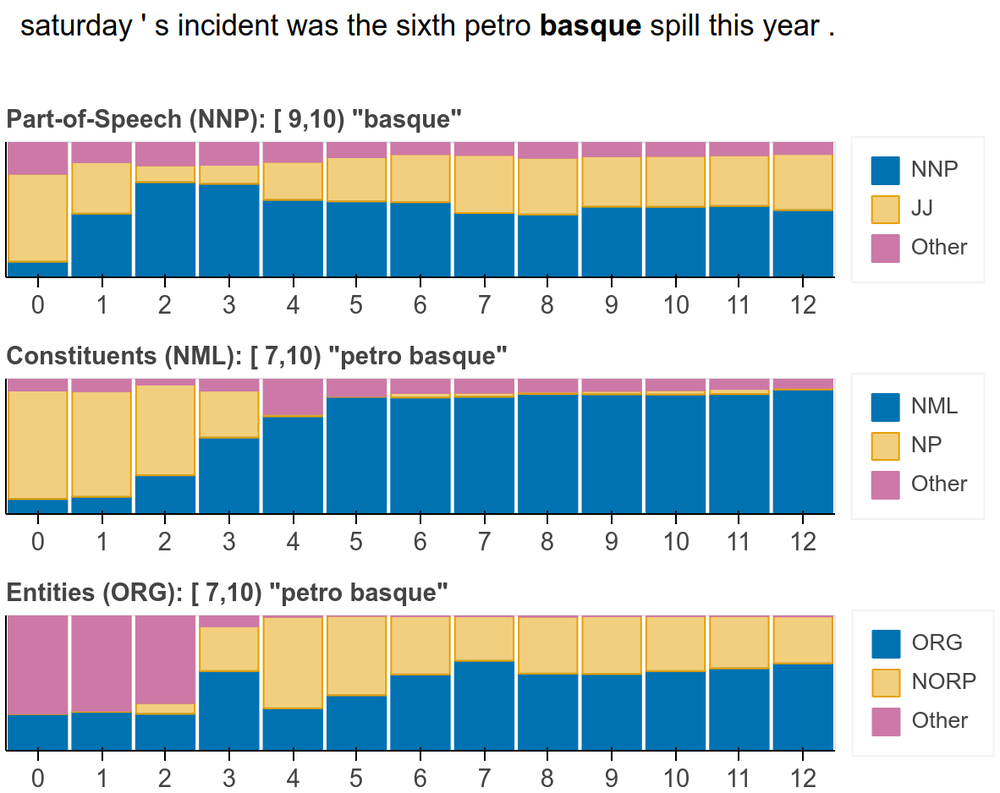}
    \caption{Trace of selected annotations that intersect the token ``\textbf{basque}'' in the above sentence. We see the model recognize this as part of a proper noun (\texttt{NNP}) in layer 2, which leads it to revise its hypothesis about the constituent ``petro basque'' from an ordinary noun phrase (\texttt{NP}) to a nominal mention (\texttt{NML}) in layers 3-4. We also see that from layer 3 onwards, the model recognizes ``petro basque'' as either an organization (\texttt{ORG}) or a national or religious group (\texttt{NORP}), but does not strongly disambiguate between the two.}
    \label{fig:supp-example-petro-basque}
\end{figure}

\begin{figure}[ht!]
    \centering
    \includegraphics[width=\linewidth]{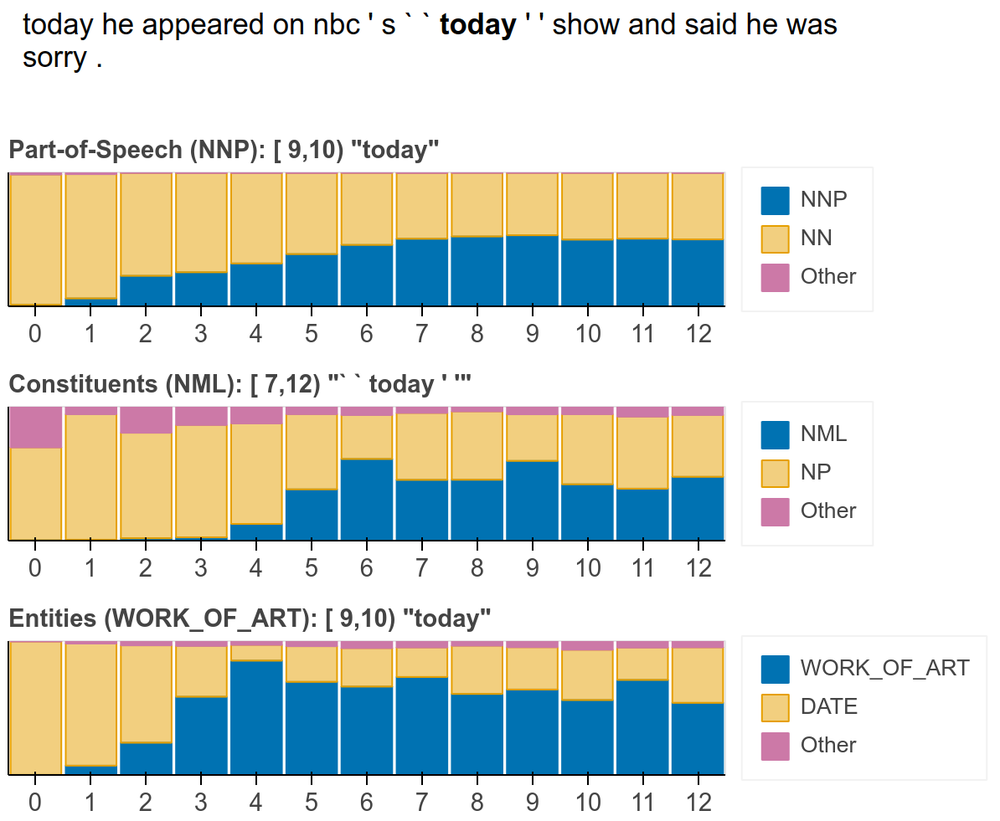}
    \caption{Trace of selected annotations that intersect the second ``\textbf{today}'' in the above sentence. The odel initially believes this to be a date and a common noun, but by layer 4 realizes that this is the TV show (entity tag \texttt{WORK\_OF\_ART}) and subsequently revises its hypotheses about the constituent type and part-of-speech.}
    \label{fig:supp-example-today-show}
\end{figure}

\begin{figure}[ht!]
    \centering
    \includegraphics[width=\linewidth]{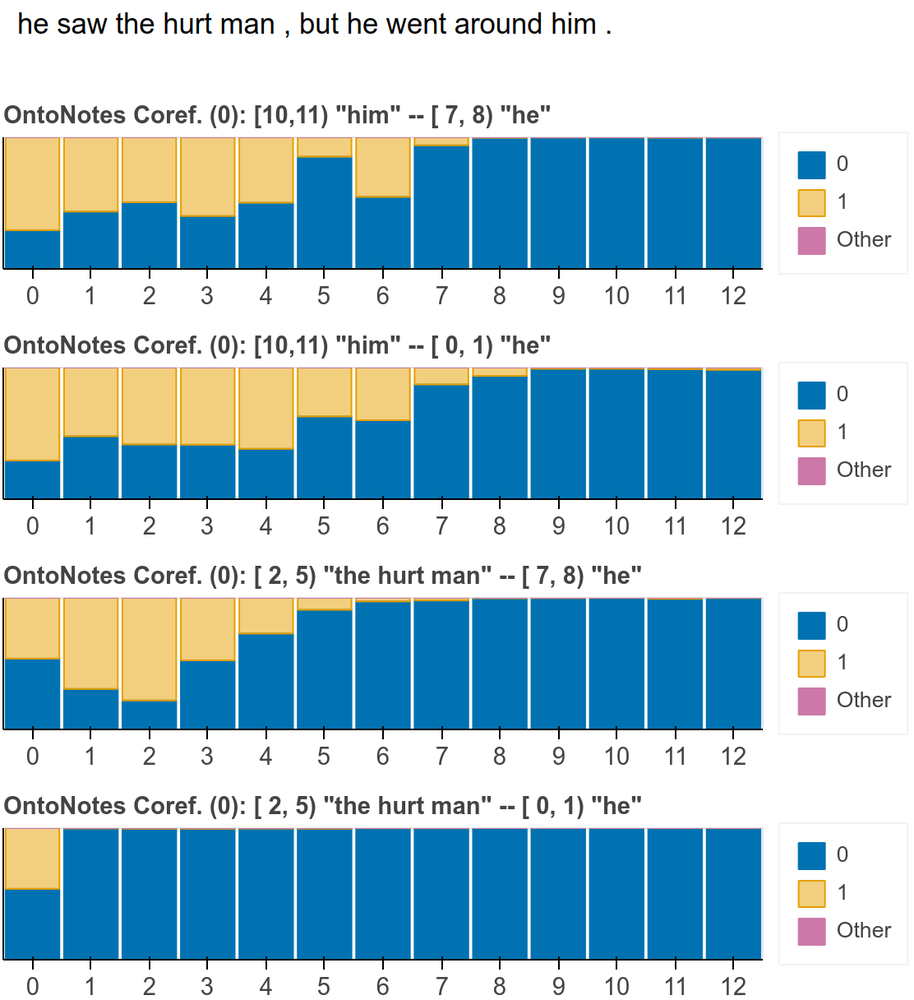}
    \caption{Trace of selected coreference annotations on the above sentence. Not shown are two coreference edges that the model has correctly resolved at layer 0 (guessing from embeddings alone): ``him'' and ``the hurt man'' are coreferent, as are ``he'' and ``he''. We see that the remaining edges, between non-coreferent mentions, are resolved in several stages.}
    \label{fig:supp-example-hurt-man}
\end{figure}

\begin{figure}[ht!]
    \centering
    \includegraphics[width=\linewidth]{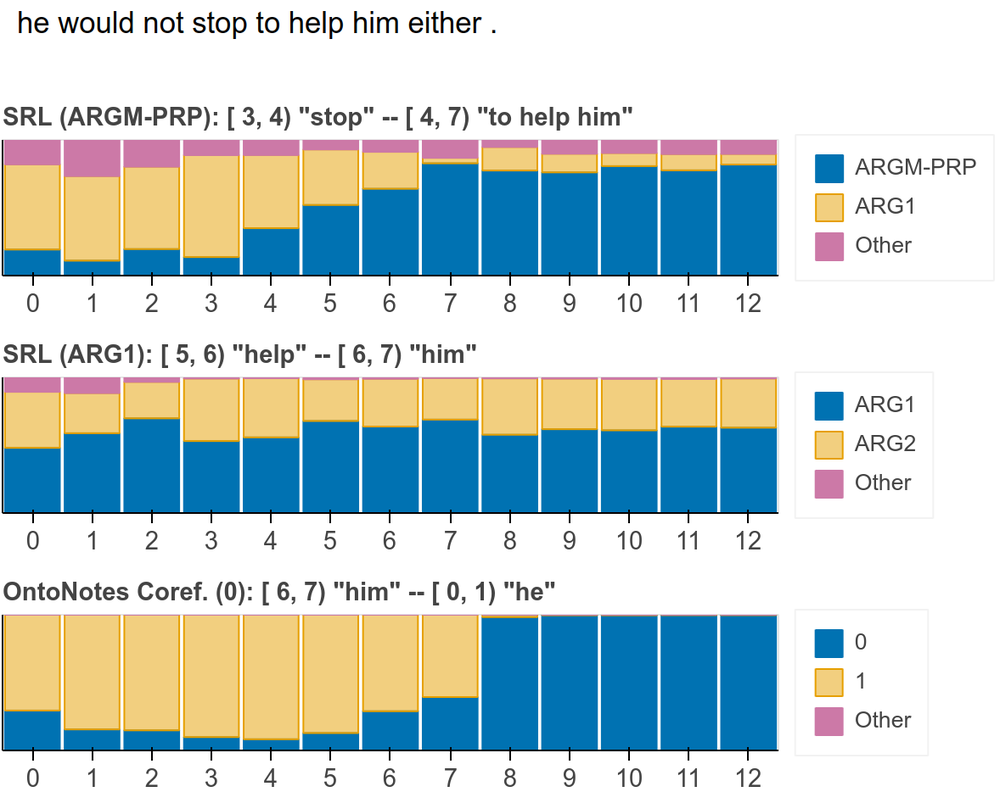}
    \caption{Trace of selected coreference and SRL annotations on the above sentence. The model resolves the semantic role (purpose, \texttt{ARGM-PRP}) of the phrase ``to help him'' in layers 5-7, then quickly resolves at layer 8 that ``him'' and ``he'' (the agent of ``stop'') are not coreferent. Also shown is the correct prediction that ``him'' is the recipient (\texttt{ARG1}, patient) of ``help''.}
    \label{fig:supp-example-help-him}
\end{figure}

\end{document}